\documentclass[letterpaper]{article} 
\usepackage{aaai2027}  
\usepackage[hyphens]{url}  
\usepackage{graphicx} 
\usepackage{natbib}  
\usepackage{caption} 
\usepackage{algorithm}
\usepackage{algorithmic}

\usepackage{newfloat}
\usepackage{listings}
\DeclareCaptionStyle{ruled}{labelfont=normalfont,labelsep=colon,strut=off} 
\floatstyle{ruled}
\newfloat{listing}{tb}{lst}{}
\floatname{listing}{Listing}

\usepackage{booktabs}

\usepackage{amsmath}
\usepackage{amssymb}
\usepackage{mathtools}
\usepackage{amsthm}
\usepackage{mathrsfs}
\usepackage{bm}
\usepackage{multirow}
\usepackage[caption=false]{subfig}
\newcommand{\safeincludegraphics}[2][]{%
  \IfFileExists{#2}{\includegraphics[#1]{#2}}{%
    \fbox{\parbox[c][1.0in][c]{1.4in}{\centering Missing figure\\\texttt{\detokenize{#2}}}}%
  }%
}

\newtheorem{theorem}{Theorem}
\newtheorem{lemma}{Lemma}
\newtheorem{assumption}{Assumption}
\newtheorem{definition}{Definition}
\title{On the Implicit Flatness Bias of Sharpness-Aware Minimization: A Linear Stability Analysis with Quantitative Hyperparameter Bounds}

\author{Jiaxin Deng\textsuperscript{\rm 1}, Junbiao Pang\textsuperscript{\rm 1}}
\affiliations{\textsuperscript{\rm 1}Beijing University of Technology \\
dengjiaxin@emails.bjut.edu.cn, junbiao\_pang@bjut.edu.cn,}

\begin{document}

\maketitle

\begin{abstract}

Sharpness-Aware Minimization (SAM) improves generalization by seeking parameters whose loss is robust to local adversarial perturbations, but the quantitative mechanism underlying its implicit bias toward flat minima remains unclear. In particular, the perturbation radius $\rho$ is typically treated as an isolated tuning parameter, despite defining the neighborhood in which SAM measures sharpness. We analyze mini-batch SAM near an interpolating minimum through linear stability. Under local linearization and gradient-noise alignment assumptions, we prove that every linearly stable minimum satisfies $\lambda_{\max}\leq\sqrt[3]{b\Gamma/(2\rho\eta^2)}$, where $\lambda_{\max}$ is the largest Hessian eigenvalue, $b$ is the batch size, $\eta$ is the learning rate, and $\Gamma$ bounds the gradient norm. The bound quantitatively characterizes SAM's implicit flatness bias: holding the other quantities fixed, a smaller batch size, a larger learning rate, or a larger radius restricts linearly stable SAM to flatter minima. It also exposes a necessary trade-off: $\rho$ should be large enough to promote flatness, yet remain local enough to preserve the approximation and stable training. We validate this prediction in a controlled study of 900 models on CIFAR-100 with ResNet-18 and VGG-19, where increasing $\rho$ is consistently associated with a smaller largest Hessian eigenvalue across batch-size and learning-rate settings. Finally, we instantiate the analysis in Taylor-Locality Controlled SAM (TLC-SAM), which adjusts $\rho$ using the observed Taylor-approximation error and further reduces the top Hessian eigenvalue relative to fixed-radius SAM. Our results provide quantitative hyperparameter bounds and a stability--locality perspective for analyzing and designing SAM variants.

\end{abstract}

\section{Introduction}
\label{sec:intro}

Generalization remains the most fundamental open challenge in deep learning, and the geometric structure of the loss landscape has long served as a core lens for understanding why overparameterized models generalize well. Sharpness-Aware Minimization (SAM)~\cite{foret-2020-SAM-ICLR} has emerged as one of the most empirically successful strategies for improving generalization by explicitly biasing optimization toward flat regions of the loss landscape. At each iteration, SAM seeks parameters whose loss remains small under a worst-case perturbation in a radius $\rho$ neighborhood. The perturbation radius $\rho$ is central to this mechanism: it sets both the neighborhood in which sharpness is measured and the strength with which SAM biases optimization toward flat regions. Despite its central role, how $\rho$ quantitatively shapes the geometry of converged minima, and how it interacts with other core optimization hyperparameters to govern generalization, remains a largely open theoretical question.

\begin{figure}[t]
    \centering
    \includegraphics[width=0.48\columnwidth]{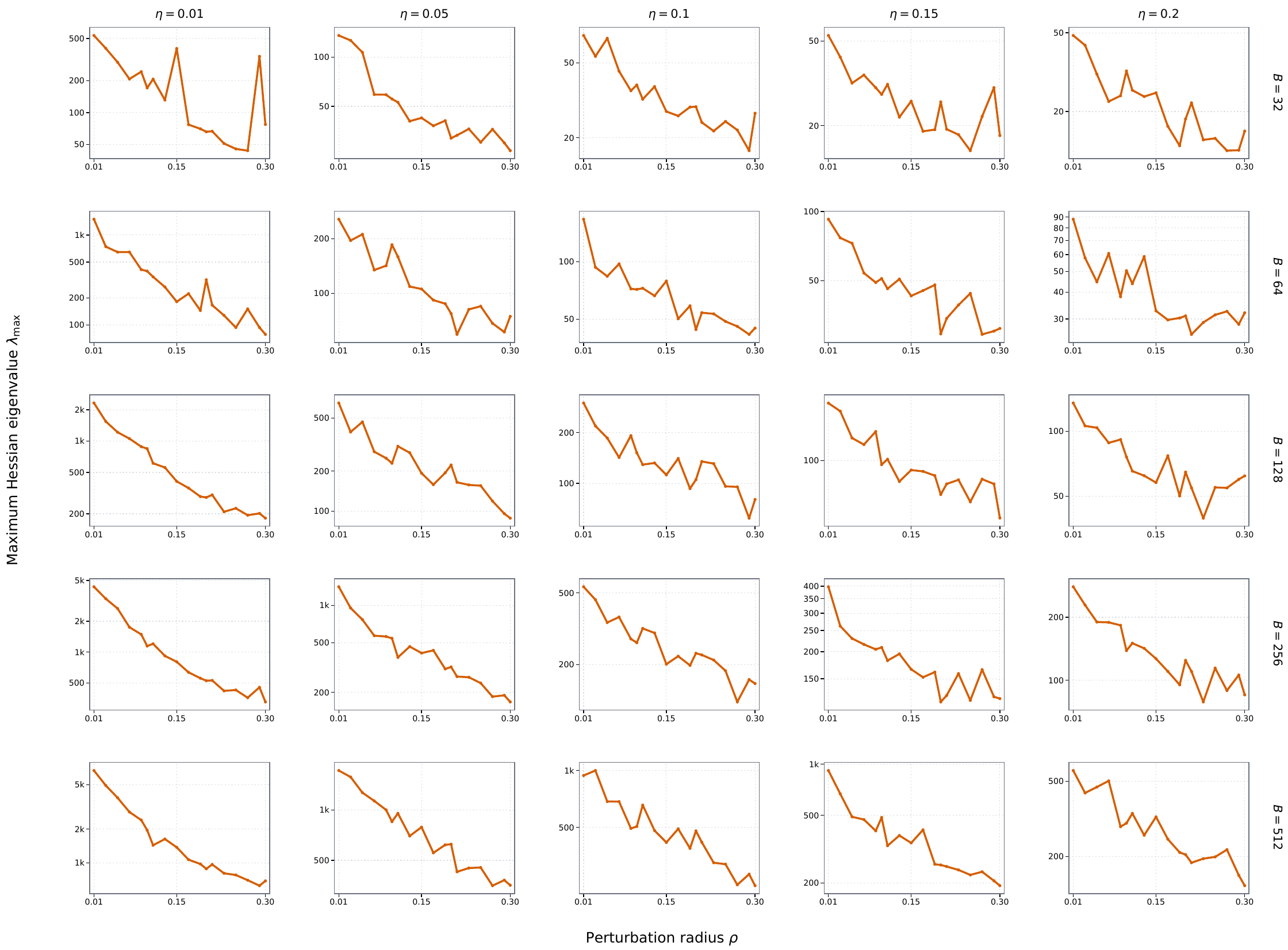}
    \hfill
    \includegraphics[width=0.48\columnwidth]{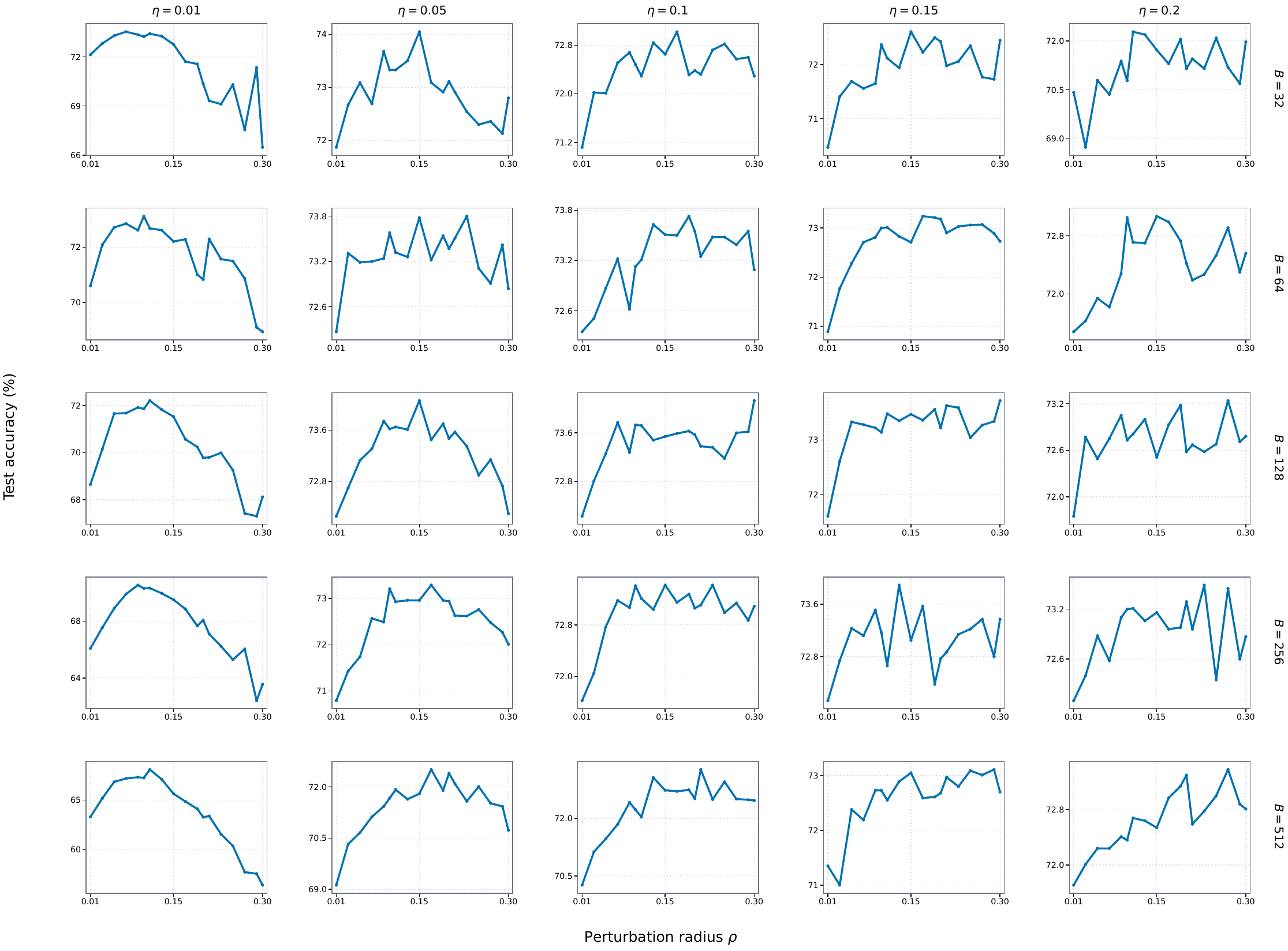}
    \caption{Effect of $\rho$ on the largest Hessian eigenvalue (left) and test accuracy (right) for ResNet-18 on CIFAR-100. Rows and columns index five batch sizes and five learning rates, respectively; each panel contains results from 18 separately trained models. Full-size panels are provided in the supplementary material.}
    \label{fig:early-rho-hessian}
\end{figure}

Recent work has made substantial progress toward explaining why SAM can generalize well. Existing theoretical investigations broadly follow three lines: (1) PAC-Bayes generalization bounds that provide initial formal justification but limited quantitative guidance for practice~\cite{andriushchenko-2022-towards_understanding-ICML,foret-2020-SAM-ICLR}; (2) analyses linking SAM to Hessian-trace regularization, which focus on convergence rather than the geometry of final minima~\cite{mulayoff2021implicit,nacsonimplicit}; and (3) analyses of learning-rate-dependent sharpness dynamics~\cite{zhousharpness,wen2022sharpness}. However, the perturbation radius and these core hyperparameters have received comparatively little dedicated study. The lack of quantitative insight into hyperparameter effects on SAM's flatness bias creates two critical gaps:

\begin{itemize}
    \item There exists no quantitative theory describing how $\rho$ couples with batch size and learning rate to jointly determine the sharpness of stable minima, making hyperparameter tuning a trial-and-error process.
    \item The lack of a mechanistic framework has led to a proliferation of SAM variants that add heuristic modifications without a principled understanding of whether such design choices are necessary or redundant.
\end{itemize}

We address these gaps by developing a unified linear-stability framework for mini-batch SAM, treating $\rho$ as an integral design variable that couples with batch size and learning rate to shape the geometry of stable minima. Adopting the linear-stability framework in \cite{wu2022alignment}, we characterize the gradient-noise covariance induced by SAM's adversarial perturbation, an effect absent in vanilla SGD. Our central theoretical result is a cubic bound on the largest Hessian eigenvalue of a linearly stable interpolating minimum: $\lambda_{\max}\leq\sqrt[3]{b\Gamma/(2\rho\eta^2)}$, where $b$ is the batch size, $\eta$ is the learning rate, $\rho$ is the perturbation radius, and $\Gamma$ is a gradient-norm bound. This bound shows that these quantities jointly constrain the sharpest curvature direction of the loss landscape. Figure~\ref{fig:early-rho-hessian} provides an early visual summary: for fixed batch size and learning rate, increasing $\rho$ generally decreases the largest Hessian eigenvalue, whereas test accuracy can deteriorate when the radius becomes overly large.

We conduct a controlled study of 900 models on CIFAR-100 with ResNet-18 and VGG-19. Across the tested batch-size and learning-rate settings, the correlation analysis supports the predicted negative relationship between $\rho$ and sharpness, while the accuracy results show the practical cost of using an overly large radius. These findings formalize a stability--locality trade-off for selecting or adapting the perturbation radius during training. Guided by this principle, we instantiate Taylor-Locality Controlled SAM (TLC-SAM), which dynamically adjusts $\rho$. TLC-SAM achieves a stronger flatness bias than fixed-radius SAM while preserving stable training.

The contributions of this paper are as follows:
\begin{itemize}
    \item We develop a linear-stability framework for mini-batch SAM and derive a cubic bound on the maximum Hessian eigenvalue of stable minima, quantifying how batch size, learning rate, and perturbation radius jointly shape loss-landscape geometry. The result identifies stability and locality as constraints on the useful perturbation radius.

    \item We present a large-scale controlled empirical study with 900 trained models, providing systematic, statistically rigorous validation of the theoretical predictions across diverse hyperparameter configurations and architectural settings.

    \item We propose TLC-SAM, a theory-guided adaptive SAM variant that dynamically adjusts perturbation radius via Taylor-locality monitoring, offering a practical instantiation of the stability-locality design principle. Our analysis further establishes a unified perspective for understanding existing SAM variants and carries implications for large model optimization.
\end{itemize}

\section{Related Works}
\label{sec:related}
\subsection{SAM and flat minima}

The relationship between model generalization and the geometry of the loss landscape has been a key focus in machine learning research. Early works, such as those by Hochreiter et al. \cite{hochreiter1994simplifying}, demonstrated a strong correlation between better generalization and flatter minima, sparking subsequent research into this connection. For instance, Keskar et al. \cite{keskar-2016-large_batch-ICLR} found that larger batch sizes increased sharpness and generalization error, while Jastrzebski et al. \cite{jastrzkebski2017three} observed a link between sharpness and the ratio of learning rate to batch size. Dinh et al. \cite{dinh-2017-sharp_minima-ICML} quantified sharpness through the eigenvalues of the Hessian matrix. However, the high computational cost associated with calculating the Hessian has hindered the practical use of this approach to identify flatter minima. Sharpness-Aware Minimization (SAM) \cite{foret-2020-SAM-ICLR} incorporates a perturbation strategy that encourages convergence to flatter regions of the loss landscape by penalizing sharpness. Theoretical frameworks relating sharpness to generalization error were explored in \cite{dziugaite2017computing}, \cite{neyshabur2017exploring}, collectively supporting the pursuit of flatter minima for improved generalization. 

\subsection{Understanding SAM}
From an optimization perspective, several works have studied the convergence and computational dynamics of SAM. \cite{andriushchenko-2022-towards_understanding-ICML,si2023practical} examine stochastic SAM and its tendency to avoid sharp minima. In contrast, \cite{compagnoni2023sde,kim2023stability} show that SAM can take longer than standard SGD to escape saddle points. \cite{long2024sharpness} further analyzes edge stability in SAM.

Regarding implicit bias, SAM is widely regarded as favoring flat minima. Using dynamical stability analysis, \cite{behdin2023msam} show that SAM encourages solutions in flatter regions of the loss landscape. The mechanisms through which SAM drives optimization toward flat minima are further studied in \cite{wen2022sharpness,compagnoni2023sde,bartlett2023dynamics}. \cite{wen2023sharpness} show that SAM's generalization benefits cannot be attributed solely to sharpness minimization, while \cite{dai2023crucial} investigate how normalization affects its optimization dynamics. Finally, \cite{zhousharpness} show that applying SAM late in training can efficiently move from sharp to flat minima.

While we build upon the linear stability framework originally developed to characterize the passive flatness bias of vanilla SGD~\cite{wu2018sgd} \cite{wu2022alignment}, SAM’s inherent active adversarial perturbation mechanism requires characterizing curvature-noise alignment, with detailed discussion in the supplementary material. Our theoretical results thus offer a unified explanation for the empirical effectiveness of representative SAM variants, including ASAM and F-SAM.

\section{Theoretical Evidence}
\subsection{Background and Preliminaries}
\textbf{Notation and Setup.}
Let $\mathcal{D}_{\rm tr} = \{({\mathbf{x}}_i, \mathbf{y}_i)\}_{i=1}^n$ and $\mathcal{D}_{\rm te} = \{({\mathbf{x}}_i, \mathbf{y}_i)\}_{i=n+1}^{n+m}$ be the training set and the testing set, where ${\mathbf{x}}_i\in \mathbb{R}^{d}$ is the input and $\mathbf{y}_i \in \mathbb{R}^{k}$ is the corresponding target. We consider a model in the form of $f(\mathbf{x}; \mathbf{w})$, where $\mathbf{w} \in \mathbb{R}^p$ are the model parameters and $f(\mathbf{x};\mathbf{w})\in\mathbb{R}^k$ is the output of the input $\mathbf{x}$.
The loss of the model for the $i$-th sample $(\mathbf{x}_i, \mathbf{y}_i)$ is denoted as $\ell(\mathbf{y}_i; f(\mathbf{x}_i; \mathbf{w}))$, simplified to $\ell_i(\mathbf{w})$.
The loss over the training set and the testing loss are then given by $\mathcal{L}_{\mathcal{D}_{\rm tr}}(\mathbf{w}) = \frac{1}{n} \sum_{i=1}^n \ell_i(\mathbf{w})$ and $\mathcal{L}_{\mathcal{D}_{\rm te}}(\mathbf{w}) = \frac{1}{m} \sum_{i=n+1}^{n+m} \ell_i(\mathbf{w})$, respectively.
Unless otherwise specified, we denote $\mathcal{L}(\mathbf{w})=\mathcal{L}_{\mathcal{D}_{\rm tr}}(\mathbf{w})$ for simplicity.

We focus on the over-parameterized case in the sense that $\min_{\mathbf{w}}\mathcal{L}(\mathbf{w})=0$.
To characterize the local geometry of the training loss at point $\mathbf{w}$, we consider the Fisher matrix
$G(\mathbf{w})=\frac{1}{n}\sum_{i=1}^n\nabla f(\mathbf{x}_i;\mathbf{w})\nabla f(\mathbf{x}_i;\mathbf{w})^{\top}$
and the Hessian matrix $H(\mathbf{w}) = G(\mathbf{w}) + \frac{1}{n}\sum_{i=1}^n ({f(\mathbf{x}_i; \mathbf{w}) - \mathbf{y}_i}) \nabla^2 f(\mathbf{x}_i; \mathbf{w})$, where $G(\mathbf{w})\approx H(\mathbf{w})$ in the low-loss region. If $\mathbf{w}^*$ is a global minimum, $G(\mathbf{w}^*) = H(\mathbf{w}^*)$. The squared loss is commonly used in theoretical analyses of implicit bias \cite{gunasekar2017implicit}, training dynamics \cite{jacot2018neural}, and convergence \cite{dugradient}. For a matrix $A$, $\|A\|_2$, $\|A\|_F$, and $\operatorname{Tr}(A)$ denote its spectral norm, Frobenius norm, and trace, respectively.

\textbf{Sharpness-Aware Minimization.} The optimization objective of SAM is defined as:
\begin{equation}\label{equ:sam_max}
\begin{aligned}
\mathop {\min } \limits_\mathbf{w} [(\mathcal{L}^{max}(\mathbf{w})&- \mathcal{L}(\mathbf{w}) ) + \mathcal{L}(\mathbf{w}) + \lambda ||\mathbf{w}||_2^2] \\ \mathcal{L}^{max}(\mathbf{w})&= \mathop {\max }\limits_{||\bm{\varepsilon} || \le \rho } \mathcal{L}(\mathbf{w} + \bm{\varepsilon} ),
\end{aligned}
\end{equation}
where $\mathbf{w}$ represents the weights of the network, $\bm{\varepsilon}$ represents the perturbation of weights $\mathbf{w}$ in a Euclidean ball with the radius $\rho$ $(\rho>0)$, $\mathcal{L}(\cdot)$ is the loss function, $\lambda\|\mathbf{w}\|_2^2$ is a $\ell_2$ regularization term. 

SAM utilizes Taylor expansion to approximate the maximized loss ($\mathcal{L}^{max}(\mathbf{w})$) in local parameter space as follows:
\begin{equation}\label{equ:psf}
\begin{aligned}
&\mathop {\arg \max }\limits_{||\bm{\varepsilon} || \le \rho } \;
\mathcal{L}(\mathbf{w} + \bm{\varepsilon} ) \approx \mathop {\arg \max }\limits_{||\bm{\varepsilon} || \le \rho } \;
\Big( \mathcal{L}(\mathbf{w})
 + {\bm{\varepsilon} ^{\top}}{\nabla _\mathbf{w}}\mathcal{L}(\mathbf{w}) \Big) \\
\end{aligned}
\end{equation}
By solving Eq.~\eqref{equ:psf}, SAM obtains the perturbation as follows:
\begin{equation}\label{equ:e_max}
\begin{aligned}
\hat{ \bm{\varepsilon} } = \rho {\nabla _\mathbf{w}}\mathcal{L}(\mathbf{w})/||{\nabla _\mathbf{w}}\mathcal{L}(\mathbf{w})||_2.
\end{aligned}
\end{equation}
Substituting the perturbation $\hat{\bm{\varepsilon}}$ back into Eq.~\eqref{equ:sam_max} and differentiating, 
we then have:
\begin{equation}\label{equ:gs}
\begin{aligned}
&{\nabla _{\mathbf{w}}}{{\mathcal{L}}^{max}}({\mathbf{w}})
\approx {\nabla _{\mathbf{w}}}\mathcal{L}({\mathbf{w}}
 + \hat {\bm{\varepsilon}} ({\mathbf{w}})) \\
&= {\nabla _{\mathbf{w}}}\mathcal{L}({\mathbf{w}})
{|_{{\mathbf{w}} + \hat {\bm{\varepsilon}} ({\mathbf{w}})}} + \frac{{d\hat {\bm{\varepsilon}} ({\mathbf{w}})}}{{d{\mathbf{w}}}}
{\nabla _{\mathbf{w}}}\mathcal{L}({\mathbf{w}})
{|_{{\mathbf{w}} + \hat {\bm{\varepsilon}} ({\mathbf{w}})}}.
\end{aligned}
\end{equation}
By dropping the second-order terms in Eq.\eqref{equ:gs}, SAM approximates the gradient as follows:
\begin{equation}\label{equ:grad_sam}
\begin{aligned}
{\nabla _{\mathbf{w}}}&{{\mathcal{L}}^{max}}({\mathbf{w}}) \approx {\nabla _\mathbf{w}}\mathcal{L}(\mathbf{w}){|_{\mathbf{w} + \bm{\hat\varepsilon} }}.
\end{aligned}
\end{equation}
SAM uses the gradient in Eq.~\eqref{equ:grad_sam} to update the weights.

\textbf{Remark.} Eq.~\eqref{equ:psf} and Eq.~\eqref{equ:gs} require $\rho$ to remain within a sufficiently local region for the first-order approximation to be valid.

To provide a theoretical analysis, we establish the following fundamental assumptions. The assumptions on smoothness and bounded variance of stochastic gradients are standard in the literature on non-convex optimization \cite{zhuang-2022-GSAM-ICLR} \cite{wu2022alignment} \cite{mori2021logarithmic}.

\begin{assumption}\label{assumpion:Bounded_g}
There exists $\Gamma>0$ such that $\mathbb{E}\!\left[\|\nabla\mathcal{L}(\mathbf{w}_t)\|_2\right]\leq\Gamma$ for every training step $t$.
\end{assumption}

\begin{assumption}\label{assumpion:1}
The full-batch gradient $\nabla_{\mathbf{w}}\mathcal{L}(\mathbf{w}_t)$ is negligible compared with the per-sample gradients $\nabla_{\mathbf{w}}\mathcal{L}_i(\mathbf{w}_t)$ \cite{wu2022alignment}.
\end{assumption}

\begin{assumption}\label{assumpion:3}
Let $\mathbf{w}^*$ be a global minimum of $\mathcal{L}(\cdot)$. The local linearization of the model around $\mathbf{w}^*$ is
\begin{equation}\label{equ:lin_model}
\begin{aligned}
f_{\text{lin}}(\mathbf{x};\mathbf{w}) = f(\mathbf{x};\mathbf{w}^*) + \langle \mathbf{w} - \mathbf{w}^*, \nabla_{\mathbf{w}} f(\mathbf{x};\mathbf{w}^*) \rangle.
\end{aligned}
\end{equation} 
\end{assumption}

The linearization is widely adopted by linear stability analysis \cite{wu2018sgd} \cite{ma2021linear}. Without loss of generality, we let $\mathbf{w}^*=0$ \cite{wu2022alignment}. For the linearized model and squared loss, we have $\mathcal{L}({\mathbf{w}_t}) = \frac{1}{2}{\mathbf{w}_t}^\top H(\mathbf{w}^*){\mathbf{w}_t}$, $\nabla \mathcal{L}({\mathbf{w}_t}) = H(\mathbf{w}^*){\mathbf{w}_t}$.

\subsection{Flatness bias of SAM}\label{sec:flatness}
By Taylor expansion, the optimization step of SAM is written as follows:
\begin{equation}\label{equ:sam_batch}
\begin{aligned}
&{\mathbf{w}_{t + 1}}
= {\mathbf{w}_t} - \eta \nabla \mathcal{L}_{\mathcal{B}_t}\!\left({\mathbf{w}_t}
 + \rho\frac{\nabla\mathcal{L}_{\mathcal{B}_t}(\mathbf{w}_t)}
 {\|\nabla\mathcal{L}_{\mathcal{B}_t}(\mathbf{w}_t)\|_2}\right) \\
&\approx {\mathbf{w}_t} - \eta \Bigg( 
\nabla \mathcal{L}_{\mathcal{B}_t}({\mathbf{w}_t}) + \rho \frac{\nabla ^2\mathcal{L}_{\mathcal{B}_t}({\mathbf{w}_t})
 \nabla \mathcal{L}_{\mathcal{B}_t}({\mathbf{w}_t})}
 {\|\nabla \mathcal{L}_{\mathcal{B}_t}({\mathbf{w}_t})\|_2} \Bigg),
\end{aligned}
\end{equation}
where $\eta$ is the learning rate, $\mathcal{B}_t$ is the mini-batch sampled at step $t$, $b=|\mathcal{B}_t|$ is its size, and $\mathcal{L}_{\mathcal{B}_t}$ is the corresponding average loss. Let $g_t=\nabla\mathcal{L}(\mathbf{w}_t)$ and $H_t=\nabla^2\mathcal{L}(\mathbf{w}_t)$. Separating the stochastic component in Eq.~\eqref{equ:sam_batch} gives:
\begin{equation}\label{equ:sam_noise_decomposition}
\begin{aligned}
{\mathbf{w}_{t + 1}}
&\approx {\mathbf{w}_t} - \eta\left(
g_t+\rho\frac{H_tg_t}{\|g_t\|_2}+\xi_t\right),
\end{aligned}
\end{equation}
where $\xi_t$ denotes the mini-batch SAM gradient noise. Under independent sampling with replacement, $\mathbb{E}[\xi_t]=0$ and $\mathbb{E}[\xi_t\xi_t^\top]=\Sigma_t/b$, where $\Sigma_t$ is the corresponding noise covariance characterized in Lemma~\ref{lemma:1}.

\begin{lemma}[]\label{lemma:1}
(Gradient-noise covariance of SAM.) Assume that $\nabla f(\mathbf{x}_i;\mathbf{w}_t)$ and $\mathcal{L}_i(\mathbf{w}_t)$ are nearly decoupled~\cite{mori2021logarithmic}. For the squared loss $\mathcal{L}_i(\mathbf{w})=\frac{1}{2}|f(\mathbf{x}_i;\mathbf{w})-\mathbf{y}_i|^2$, the covariance of $\xi_t(\mathbf{w}_t+\hat{\varepsilon}_t)$ can be approximated as follows:
\begin{equation}\label{equ:sam_noise_covariance}
\begin{aligned}
\Sigma ({\mathbf{w}_t})
&\approx 2\mathcal{L}({\mathbf{w}_t})\Bigg( G({\mathbf{w}_t})
 + \frac{{2\rho G({\mathbf{w}_t})H{{({\mathbf{w}_t})}^\top}}}
 {{||\nabla \mathcal{L}({\mathbf{w}_t})||}} \\
&\qquad
 + \frac{{{\rho ^2}H({\mathbf{w}_t})G({\mathbf{w}_t})H({\mathbf{w}_t})}}
 {{||\nabla \mathcal{L}({\mathbf{w}_t})||^2}} \Bigg)
\end{aligned}
\end{equation}
\end{lemma}

\begin{definition}[]
(Linear stability of SAM.) A global minimum $\mathbf{w}^*$ is linearly stable for SAM if there exists a constant $C>0$ such that, when SAM is applied to the linearized model in Assumption~\ref{assumpion:3}, $\mathbb{E}[\mathcal{L}(\mathbf{w}_t)]\leq C\mathbb{E}[\mathcal{L}(\mathbf{w}_0)]$ for every $t\geq0$ and every initialization $\mathbf{w}_0$ sufficiently close to $\mathbf{w}^*$ for the linearization to be valid.
\end{definition}

We adopt this definition for our analysis. \cite{wu2022alignment} demonstrated that the linear stability of stochastic algorithms is heavily influenced by gradient noise. For the mini-batch SAM, Theorem \ref{theorem:1} characterizes the relationship of $\rho$ and generalization ability.

\begin{theorem}[]\label{theorem:1}
(Sharpness bound for linearly stable SAM.) Let $\mathbf{w}^*$ be a global minimum that is linearly stable for SAM under the linearized model in Eq.~\eqref{equ:lin_model}. Under Lemma~\ref{lemma:1}, Assumptions~\ref{assumpion:Bounded_g}--\ref{assumpion:3}, and the local relation $G(\mathbf{w}_t)\approx H(\mathbf{w}^*)$, we have
\begin{equation}\label{equ:sharpness_bound}
\centering
\begin{aligned}
{\lambda _{\max }} \le \sqrt[3]{{\frac{{b\mathbb{E}\left[ {||\nabla \mathcal{L}({w_t}){\rm{||}}} \right]}}{{2\rho \eta^2}}}}\le \sqrt[3]{{\frac{{b\Gamma}}{{2\rho \eta^2}}}}
\end{aligned}
\end{equation}
where $b=|\mathcal{B}_t|$ is the batch size, $\eta$ is the learning rate, $\rho$ is the perturbation radius, and $\Gamma$ is the gradient-norm bound in Assumption~\ref{assumpion:Bounded_g}.

\end{theorem}

\textbf{Remark.} Theorem~\ref{theorem:1} shows that the largest Hessian eigenvalue is jointly bounded by the batch size $b$, learning rate $\eta$, perturbation radius $\rho$, and gradient-norm bound $\Gamma$. The largest Hessian eigenvalue is commonly used as a sharpness metric~\cite{keskar-2016-large_batch-ICLR,jastrzkebski2017three,chaudhari-2019-Entropy_sgd-IOP}. Holding the other quantities and $\Gamma$ fixed, a smaller $b$, a larger $\eta$, or a larger $\rho$ yields a smaller upper bound on $\lambda_{\max}$ and may favor better generalization, provided that the perturbation remains within the local regime of the analysis.

Despite the beneficial effects of increasing $\rho$ on the flatness of minima, $\rho$ should not be too large. The linear stability assumption requires $\rho$ to remain small enough to ensure the validity of the underlying model and analysis. A large $\rho$ may compromise the linearity assumption, which could hinder the model's optimization behavior, particularly in maintaining the stability required for SAM. 

Our proof assumes that the iterates remain close to a global minimum. Extending the analysis to the full training trajectory is challenging, as most analyses of (S)GD dynamics focus on simplified scenarios, such as single-index models~\cite{damian2024computational} or the feature-learning regime~\cite{lyu2021gradient}. Nevertheless, the local result provides guidance for selecting $\rho$ during training.

\begin{table*}[t]
    \centering
    \caption{Spearman correlation coefficients (SCC) and $p$ values between perturbation radius ($\rho$) and the largest Hessian eigenvalue on CIFAR-100 under different learning rates ($\eta$) and batch sizes ($b$).}
    \label{tab:scc-rho-hessian}
    \begingroup
    \small
    \setlength{\tabcolsep}{1mm}
    \begin{tabular}{@{}ll*{10}{c}@{}}
        \toprule
        \multirow{2}{*}{Model} & \multirow{2}{*}{$\eta$}
        & \multicolumn{2}{c}{32}
        & \multicolumn{2}{c}{64}
        & \multicolumn{2}{c}{128}
        & \multicolumn{2}{c}{256}
        & \multicolumn{2}{c}{512} \\
        \cmidrule(lr){3-4}
        \cmidrule(lr){5-6}
        \cmidrule(lr){7-8}
        \cmidrule(lr){9-10}
        \cmidrule(l){11-12}
        & & SCC  & $p$ & SCC & $p$ & SCC & $p$ & SCC & $p$ & SCC & $p$ \\
        \midrule
        \multirow{5}{*}{ResNet-18}
        & 0.01 & $-0.678$ & $2.0{\times}10^{-3}$ & $-0.950$ & $1.6{\times}10^{-9}$ & $-0.990$ & $6.3{\times}10^{-15}$ & $-0.981$ & $6.7{\times}10^{-13}$ & $-0.994$ & $1.1{\times}10^{-16}$ \\
        & 0.05 & $-0.965$ & $1.0{\times}10^{-10}$ & $-0.930$ & $2.4{\times}10^{-8}$ & $-0.950$ & $1.6{\times}10^{-9}$ & $-0.981$ & $6.7{\times}10^{-13}$ & $-0.973$ & $1.2{\times}10^{-11}$ \\
        & 0.10 & $-0.911$ & $1.5{\times}10^{-7}$ & $-0.915$ & $1.0{\times}10^{-7}$ & $-0.880$ & $1.4{\times}10^{-6}$ & $-0.926$ & $3.7{\times}10^{-8}$ & $-0.961$ & $2.5{\times}10^{-10}$ \\
        & 0.15 & $-0.765$ & $2.2{\times}10^{-4}$ & $-0.909$ & $1.7{\times}10^{-7}$ & $-0.907$ & $2.1{\times}10^{-7}$ & $-0.895$ & $5.4{\times}10^{-7}$ & $-0.963$ & $1.6{\times}10^{-10}$ \\
        & 0.20 & $-0.874$ & $2.1{\times}10^{-6}$ & $-0.783$ & $1.2{\times}10^{-4}$ & $-0.814$ & $3.9{\times}10^{-5}$ & $-0.926$ & $3.7{\times}10^{-8}$ & $-0.928$ & $3.0{\times}10^{-8}$ \\
        \midrule
        \multirow{5}{*}{VGG-19}
        & 0.01 & $-0.602$ & $8.3{\times}10^{-3}$ & $-0.629$ & $5.2{\times}10^{-3}$ & $-0.971$ & $2.2{\times}10^{-11}$ & $-0.304$ & $2.2{\times}10^{-1}$ & $-0.975$ & $6.6{\times}10^{-12}$ \\
        & 0.05 & $-0.639$ & $4.3{\times}10^{-3}$ & $-0.928$ & $3.0{\times}10^{-8}$ & $-0.971$ & $2.2{\times}10^{-11}$ & $-0.992$ & $1.1{\times}10^{-15}$ & $-0.959$ & $3.7{\times}10^{-10}$ \\
        & 0.10 & $-0.523$ & $2.6{\times}10^{-2}$ & $-0.938$ & $8.9{\times}10^{-9}$ & $-0.961$ & $2.5{\times}10^{-10}$ & $-0.938$ & $8.9{\times}10^{-9}$ & $-0.924$ & $4.6{\times}10^{-8}$ \\
        & 0.15 & $-0.963$ & $1.6{\times}10^{-10}$ & $-0.887$ & $9.7{\times}10^{-7}$ & $-0.837$ & $1.5{\times}10^{-5}$ & $-0.951$ & $1.6{\times}10^{-9}$ & $-0.934$ & $1.5{\times}10^{-8}$ \\
        & 0.20 & $-0.841$ & $1.2{\times}10^{-5}$ & $-0.515$ & $2.9{\times}10^{-2}$ & $-0.761$ & $2.5{\times}10^{-4}$ & $-0.928$ & $3.0{\times}10^{-8}$ & $-0.975$ & $6.6{\times}10^{-12}$ \\
        \bottomrule
    \end{tabular}
    \endgroup
\end{table*}

\subsection{Examples of Applying the Theory to SAM Variants}
With $b$, $\eta$, and $\Gamma$ fixed, Theorem~\ref{theorem:1} shows that a larger useful $\rho$ lowers the upper bound on $\lambda_{\max}$ within the locally stable regime. A SAM variant can therefore make a larger perturbation practical or define it in scale-aware coordinates. We illustrate these implications through ASAM and F-SAM, with the complete derivations and applicability conditions deferred to the supplementary material.

\textbf{ASAM.}
Let $\mathbf{g}=\nabla\mathcal{L}(\mathbf{w})$. ASAM~\cite{kwon-2021-asam-ICML} uses the relative neighborhood $\|T_{\mathbf{w}}^{-1}\boldsymbol{\epsilon}\|_2\leq\rho_0$, where the positive diagonal matrix $T_{\mathbf{w}}$ encodes parameter scales. Under this common locality budget, its first-order maximizer is $\boldsymbol{\epsilon}_{\mathrm{ASAM}}=\rho_0T_{\mathbf{w}}^2\mathbf{g}/\|T_{\mathbf{w}}\mathbf{g}\|_2$. Defining the gradient-direction effective radius as $\boldsymbol{\epsilon}^{\top}\mathbf{g}/\|\mathbf{g}\|_2$, Cauchy--Schwarz gives
\[
\rho_{\mathrm{eff}}^{\mathrm{ASAM}}
=\rho_0\frac{\|T_{\mathbf{w}}\mathbf{g}\|_2}{\|\mathbf{g}\|_2}
\geq\rho_0\frac{\|\mathbf{g}\|_2}{\|T_{\mathbf{w}}^{-1}\mathbf{g}\|_2}
=\rho_{\mathrm{eff}}^{\mathrm{SAM}}.
\]
Thus, ASAM can apply a stronger first-order sharpness perturbation than a standard SAM perturbation under the same relative locality constraint. This accords with Theorem~\ref{theorem:1}'s qualitative message that a larger useful perturbation strengthens the bias toward flat minima, while the scale-aware constraint prevents a single absolute radius from being too aggressive for small-scale parameters.

\textbf{F-SAM.} F-SAM~\cite{li2024friendly} estimates the full gradient with an exponential moving average and constructs the perturbation from the residual $\mathbf{g}_{\mathcal{B}}-\widehat{\mathbf{g}}$. Suppose that $\widehat{\mathbf{g}}\approx\mathbf{g}$ and write $\mathbf{g}_{\mathcal{B}}=\mathbf{g}+\boldsymbol{\xi}$, where $\boldsymbol{\xi}$ is the centered batch-specific variation. In the gradient-dominant regime $\|\boldsymbol{\xi}\|_2\ll\|\mathbf{g}\|_2$, normalization makes the batch-dependent part of the SAM perturbation proportional to $(I-\mathbf{u}\mathbf{u}^{\top})\boldsymbol{\xi}$, with magnitude $O(\rho\|\boldsymbol{\xi}\|_2/\|\mathbf{g}\|_2)$ for $\mathbf{u}=\mathbf{g}/\|\mathbf{g}\|_2$. Let $V(\boldsymbol{\epsilon})=\mathbb{E}_{\mathcal{B}}[\|\boldsymbol{\epsilon}-\mathbb{E}_{\mathcal{B}}[\boldsymbol{\epsilon}]\|_2]$ measure the batch-to-batch variation of a perturbation. In the idealized case $\widehat{\mathbf{g}}=\mathbf{g}$, with a symmetric residual distribution and a sufficiently small residual, F-SAM normalizes the residual itself and satisfies
\[
V(\boldsymbol{\epsilon}_{\mathrm{SAM}})
<V(\boldsymbol{\epsilon}_{\mathrm{F\text{-}SAM}})
=\rho.
\]
Thus, F-SAM removes the shared component aligned with descent and yields a larger batch-to-batch perturbation variation. This variation is not itself a term in Theorem~\ref{theorem:1}; however, by reducing repeated interference with the descent update, it may permit a larger radius $\rho$ to remain locally valid. When this occurs, Theorem~\ref{theorem:1} gives a smaller upper bound on $\lambda_{\max}$, which may favor better generalization.

\section{Empirical Evidence}
\subsection{Implementation Details}
To evaluate Theorem~\ref{theorem:1}, we conduct the controlled correlation study on CIFAR-100 with ResNet-18 and VGG-19. For each architecture, we train a grid of five batch sizes, five learning rates, and 18 perturbation radii for 200 epochs, yielding 900 models in total. Detailed dataset descriptions, optimization settings are provided in the supplementary material.

The collected data is then used to investigate whether the empirical correlation follows Theorem~\ref{theorem:1}. Specifically, we calculate Spearman's rank-order correlation coefficients (SCC) and the corresponding $p$ values. For each fixed combination of architecture, batch size, and learning rate, the 18 perturbation radii correspond to 18 separately trained models. Each SCC and its corresponding $p$ value in Table~\ref{tab:scc-rho-hessian} are computed from the resulting 18 paired observations of $\rho$ and the largest Hessian eigenvalue. SCC measures the strength and direction of a monotonic association between two variables based on their ranks, without requiring the association to be linear. It ranges from $-1$ to $1$: a negative SCC means that larger values of $\rho$ tend to be ranked with smaller values of the largest Hessian eigenvalue, and a magnitude closer to $1$ indicates a stronger monotonic relationship. The corresponding $p$ value tests the null hypothesis of no rank association; a small $p$ value indicates that the observed relationship is unlikely to arise from random ordering alone. Thus, for fixed $b$ and $\eta$, a significantly negative SCC directly tests the ordering predicted by Theorem~\ref{theorem:1}, although correlation by itself does not establish causality. Table~\ref{tab:scc-rho-hessian} reports the finalized statistics for both ResNet-18 and VGG-19.

\subsection{Empirical Results on the Correlation}\label{Correlation}
\textbf{Correlation between perturbation radius ($\rho$) and the largest Hessian eigenvalue.} Following the correlation analysis protocol in prior empirical studies, we fix the learning rate and batch size, and vary the perturbation radius $\rho$ over 18 values on CIFAR-100. For each setting, we record the peak test accuracy and compute the largest Hessian eigenvalue of the corresponding best model. Since a smaller largest Hessian eigenvalue indicates a flatter solution and is commonly associated with better generalization, this statistic provides direct evidence for the flatness effect predicted by our theory.

Table~\ref{tab:scc-rho-hessian} reports the SCCs and $p$ values between $\rho$ and the largest Hessian eigenvalue for both architectures. For ResNet-18, all 25 SCCs are negative, ranging from $-0.678$ to $-0.994$, and all $p$ values are below $0.005$. For VGG-19, all 25 SCCs are also negative, ranging from $-0.3044$ to $-0.9917$, and 24 of the 25 correlations are statistically significant at the $0.05$ level. The only nonsignificant case occurs at $\eta=0.01$ and $b=256$ ($\mathrm{SCC}=-0.3044$, $p=0.2193$). Inspection of the corresponding measurements shows that this weaker correlation is driven by an anomalous value in the largest-Hessian-eigenvalue estimation rather than a systematic reversal of the negative trend; we retain the original result in Table~\ref{tab:scc-rho-hessian} for transparency. Thus, across two architectures and nearly all tested learning-rate and batch-size settings, a larger $\rho$ is associated with a smaller largest Hessian eigenvalue on CIFAR-100, provided that the perturbation remains within the stable local regime. These results support Theorem~\ref{theorem:1}, which predicts that the sharpness upper bound decreases as $\rho$ increases.

\begin{table}[t]
    \centering
    \caption{Spearman correlation between peak test accuracy and the theoretical factor $\sqrt[3]{b/(2\rho\eta^2)}$ on CIFAR-100.}
    \label{tab:generalization-theory-factor}
    \begin{tabular}{lcc}
        \toprule
        Model & SCC & $p$ \\
        \midrule
        ResNet-18 & $0.905$ & $7.7{\times}10^{-105}$ \\
        VGG-19 & $0.907$ & $5.0{\times}10^{-106}$ \\
        \bottomrule
    \end{tabular}
\end{table}

\begin{table*}[t]
    \centering
    \caption{Reported test accuracy (\%) on CIFAR-10 (C10) and CIFAR-100 (C100) with ResNet-18, WideResNet-28-10 (WRN-28-10), and PyramidNet-110 (Pyr.-110).}
    \label{tab:tlc-sam-results}
    \begingroup
    \small
    \setlength{\tabcolsep}{5pt}
    \begin{tabular}{@{}l*{6}{c}@{}}
        \toprule
        \multirow{2}{*}{Method}
        & \multicolumn{2}{c}{ResNet-18}
        & \multicolumn{2}{c}{WRN-28-10}
        & \multicolumn{2}{c}{Pyr.-110} \\
        \cmidrule(lr){2-3}
        \cmidrule(lr){4-5}
        \cmidrule(l){6-7}
        & C10 & C100 & C10 & C100 & C10 & C100 \\
        \midrule
        SGD
        & $96.18{\pm}0.09$ & $79.89{\pm}0.38$
        & $96.93{\pm}0.05$ & $82.56{\pm}0.27$
        & $97.10{\pm}0.08$ & $83.38{\pm}0.21$ \\
        SAM~\cite{foret-2020-SAM-ICLR}
        & $96.74{\pm}0.05$ & $81.08{\pm}0.27$
        & $97.52{\pm}0.05$ & $84.71{\pm}0.21$
        & $97.65{\pm}0.06$ & $86.06{\pm}0.16$ \\
        \midrule
        ASAM~\cite{kwon-2021-asam-ICML}
        & $96.63{\pm}0.15$ & $81.68{\pm}0.12$
        & $97.63{\pm}0.13$ & $84.99{\pm}0.22$
        & $97.82{\pm}0.07^{\dagger}$ & $86.47{\pm}0.09^{\dagger}$ \\
        FisherSAM~\cite{kim2022fisher}
        & $96.72{\pm}0.03$ & $80.99{\pm}0.13$
        & $97.46{\pm}0.18$ & $84.91{\pm}0.07$
        & $97.64{\pm}0.09^{\dagger}$ & $86.53{\pm}0.07^{\dagger}$ \\
        SSAM-D~\cite{mi2022make}
        & $96.87{\pm}0.04$ & $80.58{\pm}0.44$
        & $97.72{\pm}0.24$ & $84.99{\pm}0.79$
        & -- & -- \\
        F-SAM~\cite{li2024friendly}
        & $96.75{\pm}0.09$ & $81.29{\pm}0.12$
        & $97.53{\pm}0.11$ & $85.16{\pm}0.07$
        & $97.84{\pm}0.05^{\dagger}$ & $86.70{\pm}0.14^{\dagger}$ \\
        Eigen-SAM~\cite{luo2024explicit}
        & $95.9{\pm}0.2$ & $78.3{\pm}0.2$
        & $96.8{\pm}0.1$ & $82.8{\pm}0.1$
        & -- & -- \\
        Unified VaSSO~\cite{oikonomousharpness}
        & $96.34{\pm}0.01$ & $80.01{\pm}0.07$
        & $97.06{\pm}0.09$ & $83.66{\pm}0.19$
        & -- & -- \\
        \midrule
        \textbf{TLC-SAM} & $96.76{\pm}0.14$ & $81.61{\pm}0.18$ &$97.51{\pm}0.02$ &$84.92{\pm}0.14$ &$97.78{\pm}0.01$ &$86.09{\pm}0.15$ \\
        \bottomrule
    \end{tabular}
    \endgroup
    \par\smallskip
    \raggedright\scriptsize
    SGD, SAM, and TLC-SAM are evaluated under our implementation. For ASAM, FisherSAM, and F-SAM, we report the results in \cite{li2024friendly}; the remaining comparison results are reported from their respective original papers. $^{\dagger}$ Results obtained with 300 training epochs; ``--'' denotes an unreported result.
\end{table*}

\textbf{Correlation between generalization ability and $\sqrt[3]{b/(2\rho\eta^2)}$.} Theorem~\ref{theorem:1} identifies $\sqrt[3]{b/(2\rho\eta^2)}$ as the hyperparameter-dependent factor in the upper bound on $\lambda_{\max}$ when the gradient-norm bound $\Gamma$ is fixed. We therefore examine whether this quantity is also associated with generalization ability. Rather than fixing the batch size and learning rate, we pool the models trained with all batch-size, learning-rate, and perturbation-radius combinations. For each architecture, these configurations yield 279 distinct values of $\sqrt[3]{b/(2\rho\eta^2)}$. We calculate the SCC and its corresponding $p$ value between this quantity and peak test accuracy. Table~\ref{tab:generalization-theory-factor} reports the results. Both ResNet-18 and VGG-19 exhibit a strong positive rank association, with SCCs of $0.905$ and $0.907$, respectively, and extremely small $p$ values. Thus, across the pooled hyperparameter configurations, $\sqrt[3]{b/(2\rho\eta^2)}$ is strongly associated with peak test accuracy for both architectures. Since this analysis varies $b$, $\eta$, and $\rho$ simultaneously, it characterizes their aggregate association with generalization rather than the isolated effect of any one hyperparameter.

\section{An Example Method Derived from the Theoretical Results: TLC-SAM}
Theorem~\ref{theorem:1} suggests that the perturbation radius should not be treated as an isolated constant. For fixed batch size and learning rate, increasing $\rho$ strengthens the flatness bias because it decreases the upper bound of $\lambda_{\max}$. However, the proof also relies on the local Taylor approximation used by SAM. Therefore, a useful practical rule is not simply to choose the largest possible $\rho$, but to keep $\rho$ as large as the local approximation allows.

Based on this principle, we give a concrete example method, called Taylor-Locality Controlled SAM (TLC-SAM). At each iteration, SAM constructs a perturbation $\hat{\bm{\varepsilon}}_t$ with radius $\rho_t$ along the normalized gradient direction and evaluates whether the resulting loss variation is well approximated by the first-order Taylor expansion. Specifically, let
\begin{equation}
\begin{aligned}
\Delta_t^{\mathrm{lin}}
=\nabla\mathcal{L}(\mathbf{w}_t)^\top\hat{\bm{\varepsilon}}_t, \;
\Delta_t^{\mathrm{act}}
=\mathcal{L}(\mathbf{w}_t+\hat{\bm{\varepsilon}}_t)
-\mathcal{L}(\mathbf{w}_t),
\end{aligned}
\end{equation}
where $\Delta_t^{\mathrm{lin}}$ and $\Delta_t^{\mathrm{act}}$ denote the predicted first-order loss change and the actual perturbed loss change, respectively. 
We define the normalized Taylor-locality error as
\begin{equation}
\begin{aligned}
\mathcal{E}_t
=\left|\Delta_t^{\mathrm{act}}-\Delta_t^{\mathrm{lin}}\right|
/\left(\left|\Delta_t^{\mathrm{lin}}\right|+\epsilon\right),
\end{aligned}
\end{equation}
where $\epsilon$ is a small numerical constant. A small $\mathcal{E}_t$ means that the perturbation remains local and the linear approximation is reliable; a large $\mathcal{E}_t$ indicates that $\rho_t$ may be too aggressive for the current region.

To avoid reacting to a noisy single mini-batch, the method tracks an exponential moving average
\begin{equation}
\begin{aligned}
\bar{\mathcal{E}}_t
=\beta\bar{\mathcal{E}}_{t-1}
+(1-\beta)\mathcal{E}_t.
\end{aligned}
\end{equation}
After a warm-up stage, $\rho$ is updated only once every $K$ steps by comparing the current smoothed Taylor error with the value at the previous update:
\begin{equation}
\begin{aligned}
r_t
={\bar{\mathcal{E}}_t}/(
{\bar{\mathcal{E}}_{\mathrm{prev}}+\epsilon}).
\end{aligned}
\end{equation}
The update multiplier is
\begin{equation}
\begin{aligned}
m_t
&=\exp\left(
-\frac{\log r_t}{1+|\log r_t|}
\right),\\
\rho_{t+1}
&=\operatorname{clip}(\rho_t m_t,\rho_0,\rho_{\max}).
\end{aligned}
\end{equation}

We use the exponential update for three reasons. Writing $u_t=\log r_t$ gives $\log(\rho_{t+1}/\rho_t)=-u_t/(1+|u_t|)$ before clipping. First, since $m_t>0$, the update always preserves $\rho_t>0$ and responds to the relative, rather than absolute, change in Taylor error. Second, reciprocal error ratios $r_t$ and $1/r_t$ yield reciprocal multipliers $m_t$ and $1/m_t$, treating comparable error increases and decreases symmetrically on a logarithmic scale. Finally, because $-u_t/(1+|u_t|)\in(-1,1)$, the multiplier satisfies $m_t\in(e^{-1},e)$, so no single noisy error ratio can change $\rho_t$ by an unbounded factor. When the Taylor error decreases, $r_t<1$ and $m_t>1$, so $\rho$ increases to strengthen the flatness bias. When the error increases, $r_t>1$ and $m_t<1$, so $\rho$ decreases to restore locality and stability.

This example method turns the theoretical conclusion into a concrete training rule. The perturbation radius is fixed during the early unstable stage, then becomes adaptive: it grows when SAM's Taylor model becomes more accurate and shrinks when the perturbation leaves the local regime. Thus, $\rho$ is not assumed to have a universal optimal fixed value; instead, it is treated as a dynamic quantity controlled by the trade-off between flatness-seeking strength and Taylor-locality validity.

\begin{figure*}[t]
    \centering
    \subfloat[{SGD}]{
        \includegraphics[width=0.3\textwidth]{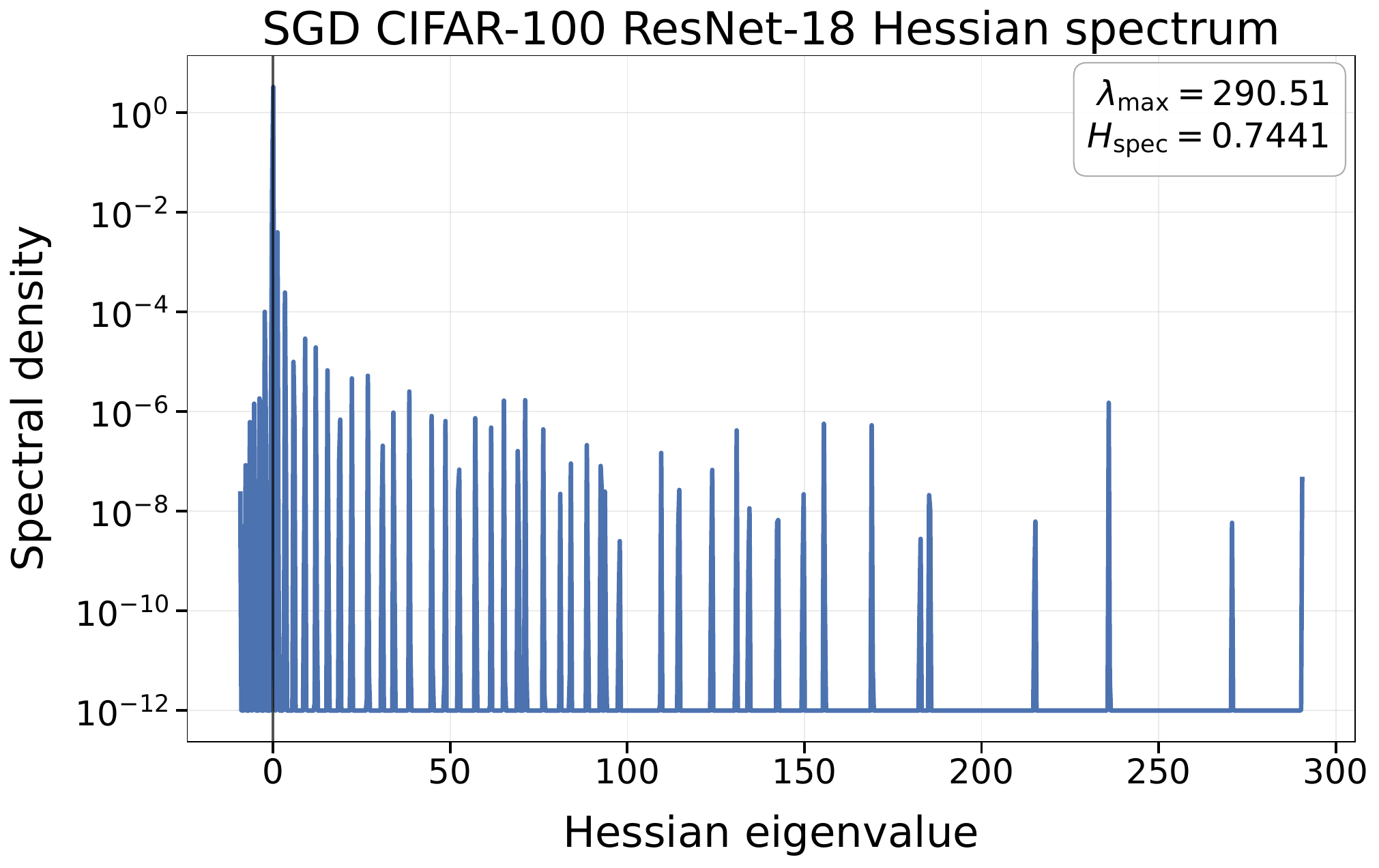}
    }
    \hfill
    \subfloat[{SAM($\rho=$)}]{
        \includegraphics[width=0.3\textwidth]{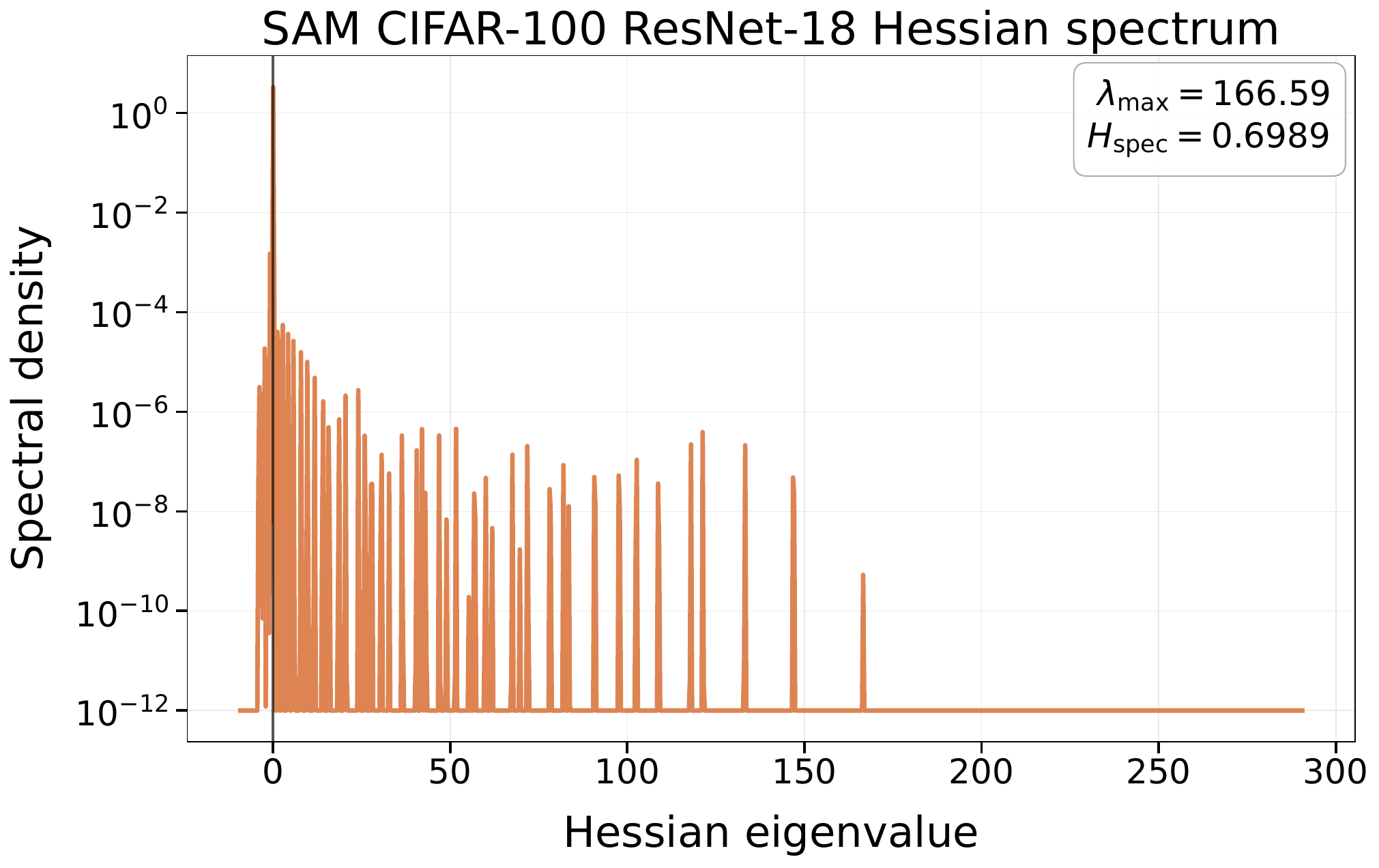}
    }
    \hfill
    \subfloat[{TLC-SAM($\rho=$)}]{
        \includegraphics[width=0.3\textwidth]{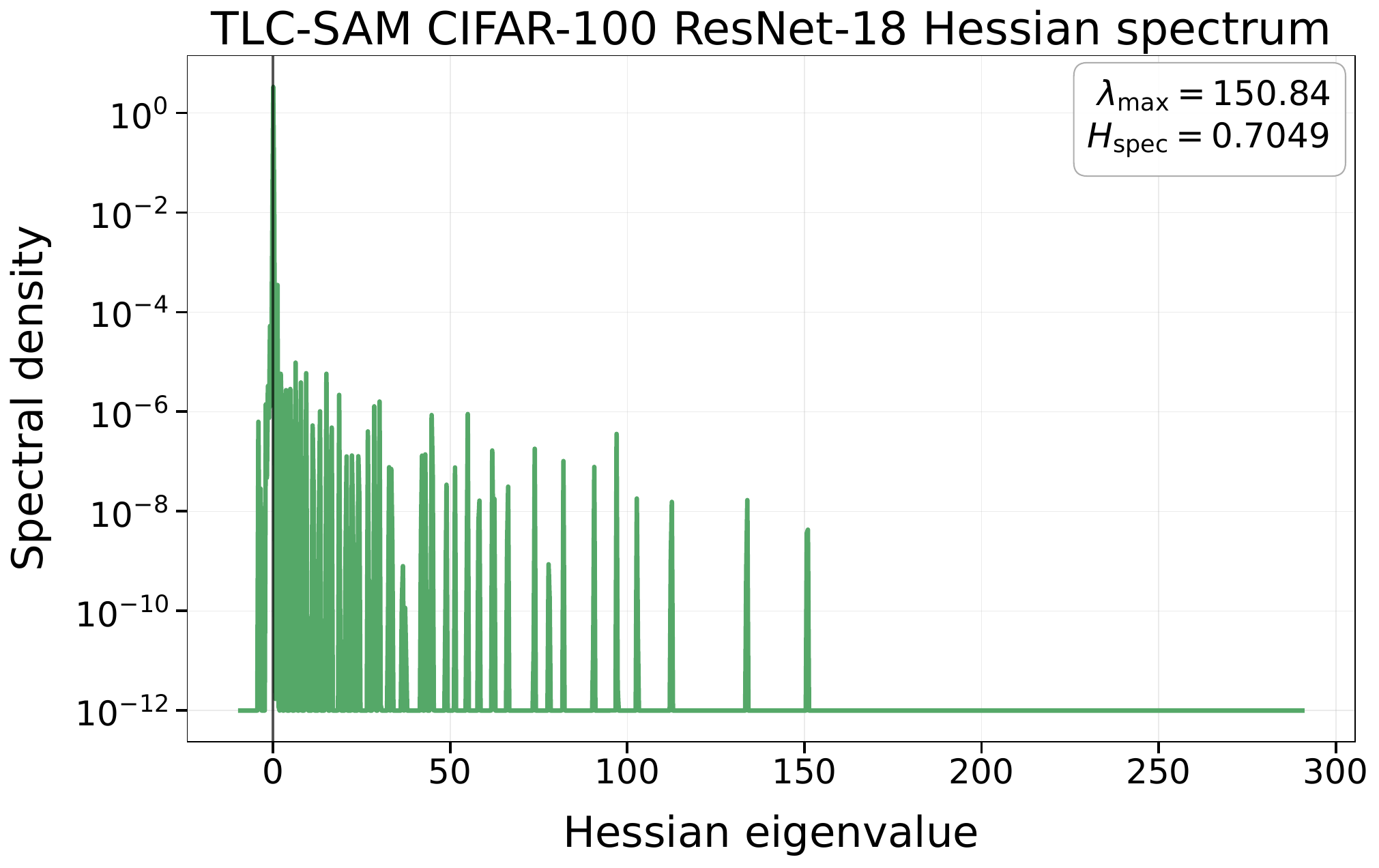}
    }
    \caption{Hessian spectral densities of ResNet-18 trained on CIFAR-100 with SGD, SAM, and TLC-SAM. The dashed vertical line in each panel marks the corresponding largest Hessian eigenvalue.}
    \label{fig:tlcsam-resnet-hessian-spectrum}
\end{figure*}

\subsection{Experimental Evaluation}
To evaluate whether the simple Taylor-locality controller improves fixed-radius SAM, we test TLC-SAM on CIFAR-10 and CIFAR-100~\cite{krizhevsky2009learning} with ResNet-18~\cite{he2016deep}, WideResNet-28-10~\cite{zagoruyko2016wide}, and PyramidNet-110~\cite{han2017deep}. The complete training protocol, TLC-SAM hyperparameters are provided in the supplementary material.  Table~\ref{tab:tlc-sam-results} reports the test accuracies.

On ResNet-18, TLC-SAM improves over SAM by $0.02$ and $0.53$ percentage points on CIFAR-10 and CIFAR-100, respectively. On WideResNet-28-10, it is essentially tied on CIFAR-10 ($-0.01$ points) and improves CIFAR-100 accuracy by $0.21$ points. On PyramidNet-110, it improves SAM by $0.13$ and $0.03$ points on CIFAR-10 and CIFAR-100. Thus, TLC-SAM improves SAM in five of the six settings and differs by only $0.01$ percentage points in the remaining setting, showing that the simple Taylor-locality controller generally improves SAM's test performance. Its results also fall in the range of the dedicated SAM variants reported in the literature. Because these variant results come from different papers and some PyramidNet-110 results use 300 training epochs, the table is not intended as a strict ranking among methods. In addition to test accuracy, we record $\rho_t$ and $\bar{\mathcal{E}}_t$ to examine whether the controller enlarges the perturbation when the local approximation becomes more reliable and contracts it when locality deteriorates. For TLC-SAM on ResNet-18/CIFAR-100, $\rho_t$ is fixed at $0.1$ for 10 epochs, stays near $0.12$--$0.15$, and after about epoch 145 often reaches $0.3$ (see the supplementary material). This suggests that TLC-SAM uses a larger radius only when locality permits, thereby strengthening its flatness bias and potentially improving generalization.

Eigen-SAM~\cite{luo2024explicit} explicitly aligns its perturbation with the leading Hessian eigen-direction to reduce $\lambda_{\max}$. Our analysis provides a complementary perspective: the gradient-noise--curvature coupling of SAM already biases the dynamics against large curvature, while Theorem~\ref{theorem:1} shows that a locally valid larger radius can further tighten the permitted upper bound on $\lambda_{\max}$. We therefore apply the Taylor-locality controller to Eigen-SAM, retaining its eigen-direction alignment while replacing its fixed radius with $\rho_t$. As reported in the supplementary material, TLC-controlled Eigen-SAM improves the test accuracy from $78.10\%$ to $78.52\%$ and lowers $\lambda_{\max}$ from $459.93$ to $428.34$. These results suggest that locality-aware radius adaptation may be more effective than optimizing the top eigenvalue alone: it can complement explicit eigen-direction regularization while allowing the perturbation scale to adapt to the local geometry.

\subsection{Effects on Eigenvalues of Hessian}

\textbf{Top Eigenvalue.} We employ Hutchinson’s method \cite{yao2020pyhessian} to compute the eigenvalues. As shown in Fig.~\ref{fig:tlcsam-resnet-hessian-spectrum}, the largest eigenvalue decreases from $290.51$ for SGD to $166.59$ for SAM, a reduction of $42.7\%$. TLC-SAM further reduces it to $150.84$, which is $9.5\%$ lower than SAM and $48.1\%$ lower than SGD. Thus, SAM substantially suppresses the sharpest direction of the loss landscape, while TLC-SAM provides an additional flattening effect. The smaller dominant curvature indicates that TLC-SAM is less sensitive to parameter perturbations along the sharpest Hessian direction, consistent with the prediction of Theorem~\ref{theorem:1}.

\textbf{Spectral Dispersion.} We compute the spectral entropy $H_{\mathrm{spec}}$ to characterize how the Hessian spectral mass is distributed across curvature regions. We integrate the normalized spectral density over consecutive eigenvalue intervals using the trapezoidal rule to obtain probabilities $p_j$, and define $H_{\mathrm{spec}}=-\sum_j p_j\log p_j$. A larger $H_{\mathrm{spec}}$ indicates a more dispersed and balanced eigenvalue distribution, whereas a smaller value indicates that the spectral mass is concentrated in fewer curvature regions. The spectral entropy decreases from $0.7441$ for SGD to $0.6989$ for SAM. TLC-SAM has an entropy of $0.7049$, which remains lower than that of SGD but is slightly higher than that of SAM. In contrast, SGD has a much larger top eigenvalue, indicating a substantially sharper dominant direction despite its higher entropy. The modest entropy increase over SAM suggests that TLC-SAM distributes the remaining spectral mass more evenly across curvature regions and may retain curvature information from a broader portion of the spectrum, while still attaining the smallest top eigenvalue.

\section{Conclusions}
This work establishes a quantitative linear-stability framework for understanding the implicit flatness bias of SAM. We prove that the largest Hessian eigenvalue of a linearly stable SAM solution obeys a cubic bound jointly determined by the batch size, learning rate, perturbation radius, and gradient-norm bound. This identifies maximum-eigenvalue suppression as a concrete mechanism through which SAM restricts the sharpest local curvature direction. It also yields a practical design principle: $\rho$ should be made as large as locality permits, rather than selected as a universal fixed constant or increased without constraint.

Controlled experiments on CIFAR-100 with ResNet-18 and VGG-19 consistently support the predicted negative association between $\rho$ and the largest Hessian eigenvalue. The spectral analysis further shows why the top eigenvalue and the remaining spectrum should be considered together: TLC-SAM attains the smallest dominant curvature while preserving a more dispersed spectral distribution than fixed-radius SAM. Thus, the contribution is not merely a tuning rule for $\rho$, but a quantitative account of how perturbation radius shapes the local geometry reached by SAM.

The framework also provides a useful basis for designing SAM variants without explicitly computing leading Hessian eigenvectors. It explains how F-SAM can make a stronger perturbation practically usable, how ASAM gives the radius a scale-aware meaning, and how TLC-SAM instantiates locality-controlled radius adaptation. Across CIFAR-10 and CIFAR-100, this simple controller generally improves over fixed-radius SAM and remains competitive with specialized variants. More broadly, the stability--locality perspective turns $\rho$ from a static hyperparameter into a controllable geometric mechanism, offering a direction for designing optimizers that jointly target generalization and stable training. Our analysis is local and relies on linearization, near-interpolating minima, and squared loss; extending it to non-interpolating regimes, highly nonlinear networks, and other loss functions remains an important direction for future work.

\bibliography{aaai2027}
\end{document}